\documentclass{article}

\usepackage{arxiv}

\usepackage[utf8]{inputenc} 
\usepackage[T1]{fontenc}    
\usepackage{hyperref}       
\usepackage{url}            
\usepackage{booktabs}       
\usepackage{tabularx}
\usepackage{threeparttable}
\usepackage{array}
\usepackage{amsmath}
\usepackage{geometry}
\usepackage{amsfonts}       
\usepackage{bbm}
\usepackage{nicefrac}       
\usepackage{microtype}      
\usepackage{titling}
\usepackage{graphicx}
\usepackage{natbib}
\usepackage{doi}
\usepackage{dirtytalk}
\usepackage{caption}
\usepackage{float}
\usepackage{subfiles}

\title{Defining AI Agents: A Compendium of Criteria, Metrics, and Benchmarks}

\author{ 
    Mia Lassiter\\
	Duke University\\
	\texttt{mia.lassiter@duke.edu}
	\and
	Brinnae Bent\\
	Duke University\\
	\texttt{brinnae.bent@duke.edu}
}

\date{}

\newcommand{\headeright}{Defining AI Agents}

\begin{document}
\maketitle

\begin{abstract}
    
    The term \say{agent} in artificial intelligence lacks a standard definition, complicating the evaluation, comparison, and reproducibility of AI agent research. We address this ambiguity through a survey organized around five dimensions of agenticness: environmental interaction, learning and adaptation, autonomy, goal-directed behavior, and temporal coherence. For each dimension, we examine how the underlying capability has been conceptualized across prior work and synthesize the metrics, benchmarks, and evaluation frameworks used to assess it. This review provides a structured account of the current landscape of agent evaluation, highlighting both established approaches and areas where evaluation remains limited or inconsistent. We additionally introduce the Agent Compendium, a public-facing digital resource that organizes and extends the evaluation methods identified through this review. Together, the survey and compendium provide a common structure for evaluating and comparing agent capabilities across AI systems, supporting more reproducible research, clearer communication, and more systematic study of artificial agents.

\end{abstract}

\section{Introduction}

Artificial intelligence (AI) agents have attracted substantial public, commercial, and academic attention because of their potential to perform complex tasks, interact with digital environments, and augment human work across domains. Yet the term \say{AI agent} lacks a widely accepted definition, and researchers continue to disagree about the capabilities required for a system to qualify as an agent. A 2026 textbook, for example, describes AI agents as \say{intelligent software entities} capable of environmental perception, autonomous decision-making, and network optimization (Mehta, 2026). A research article published that same year characterizes them as autonomous systems that complete specific tasks and communicate with other agents (Jhi and Park, 2026). These differences reflect a much longer-standing conceptual problem: researchers have used the term agent in disparate ways since the genesis of the term in computer science in the 1970s (Bent, 2025).
This ambiguity creates practical problems for AI research and deployment. First, it complicates the evaluation and reproducibility of agent research (Bent, 2025). Without shared criteria for what constitutes an agent, researchers may evaluate systems that differ substantially in their underlying capabilities while describing them using the same terminology. The rapid integration of large language models (LLMs) into agent architectures has intensified this problem, creating uncertainty about which capabilities should be evaluated and which benchmarks appropriately measure them (Zhu et al., 2025). A system that invokes tools, a system that independently pursues a multi-step objective, and a system that adapts its behavior over time may all be described as an \say{agent,} even though these systems present different evaluation requirements.

Second, inconsistent definitions complicate communication between researchers, developers, and the public (Bent, 2025). Even leading AI companies emphasize different properties when describing agents. OpenAI defines agents as \say{systems that independently accomplish tasks on your behalf} (OpenAI, 2026), whereas Anthropic describes them as 
 \say{systems where LLMs dynamically direct their own processes and tool usage, maintaining control over how they accomplish tasks} (Anthropic, 2026). Such variation shapes how users understand what agents can do, how much autonomy they possess, and what forms of human involvement they require. Shome et al. (2026), for example, found substantial mismatches between agent behavior and users’ mental models. Participants expressed uncertainty about agent capabilities, prompting requirements, and their own role during interaction. Some misunderstood when human takeover was necessary and developed what the authors describe as \say{prompt gambling}: hesitation to issue prompts because of uncertainty about how extensively an agent might act or respond (Shome et al., 2026). Definitional ambiguity therefore extends beyond terminology; it influences expectations, interaction strategies, and trust.

Third, ambiguity surrounding \say{AI agent} contributes to the broader conflation of agents with agentic AI. These terms increasingly appear across academic research, industry discourse, and public communication, often without clear distinctions between a system that qualifies as an agent and a system that exhibits particular degrees or forms of agentic behavior. Establishing explicit criteria for AI agents provides a foundation for studying agenticness as a set of measurable capabilities rather than treating \say{agent} as a catch-all label for increasingly autonomous AI systems.

To address these challenges, we introduce the Agent Compendium, a public-facing digital resource that organizes evaluation methods around the capabilities that constitute AI agency. The Agent Compendium provides explicit criteria for identifying AI agents and maps those criteria to existing metrics, benchmarks, and evaluation frameworks. By establishing clearer terminology and connecting agent criteria to concrete evaluation methods, the Agent Compendium aims to improve the reproducibility and comparability of AI agent research, support clearer communication about agent capabilities, better align user expectations with system behavior, and provide a foundation for future work on the broader concept of agentic AI.

\section{Related Work}

\subsection{Agenticness Framework}
Our research is grounded in prior work that redefines AI agents and compares AI systems’ levels of \emph{agenticness}. Agenticness is \say{the degree to which a system exhibits characteristics commonly associated with an agent} (Bent, 2025). This framework spans five dimensions: 1) environmental interaction, 2) learning and adaptation, 3) autonomy, 4) goal-directed behavior, and 5) temporal coherence. Environmental interaction refers to an agent’s understanding of and impact on its environment. Learning and adaptation refer to an agent’s ability to improve performance and modify behavior. Autonomy is an agent’s capacity to operate without external guidance. Goal-directed behavior refers to an agent’s ability to form, understand, and achieve objectives (Bent, 2025). Finally, temporal coherence describes a system’s ability to maintain logical consistency across memory, context, and intent (McCants, 2026). 

\subsection{Existing Agent Evaluations}
In 2025, Zhu et al. published a survey of LLM-based AI agent evaluations. They categorize agent evaluations based on the agents’ environments: code-related (Chen et al., 2021), web (Shi et al., 2017), OS (Xie et al., 2024), mobile (Rawles et al., 2023), scientific (Lee et al., 2023), or game (Paglieri et al., 2025). They provide a second categorization based on agent capabilities: planning (Valmeekam et al., 2023), self-reflection (Li et al., 2024), interaction (Yao et al., 2024), memory (Kočiský et al., 2018), and general (Kapoor et al., 2025; Yin et al., 2025).

In 2025, Yin et al. developed the Massive Multitask Agent Understanding (MMAU) benchmark  to evaluate agent skill sets that result in task completion or failure. Unlike existing benchmarks that merely evaluate agents’ application scenarios, the MMAU also measures agents’ abilities to understand, reason, plan, problem-solve, and self-correct across various domains. Though the MMAU benchmark is detailed, it does not evaluate AI agents’performance in interactive environments, which are critical domains (Yin et al., 2025). 

That same year, Kapoor et al. (2025) proposed the Holistic Agent Leaderboard (HAL) as \say{the missing infrastructure for AI agent evaluation.} HAL was designed to address infrastructure-, measurement-, and validation-related challenges that are unique to agent evaluation. Firstly, existing infrastructure is prohibitively slow and complex, and leaderboards are rarely updated. Secondly, agent costs deviate across token use and scaffolding, yet existing evaluations fail to account for these variations. Thirdly, agents often exploit shortcuts to optimize benchmark scores, and current evaluations seldom detect or penalize these behaviors. HAL’s three main components---a unified evaluation harness, a regularly updated, multidimensional leaderboard, and an automated analysis of agent logs---address the aforementioned challenges, respectively. HAL’s framework-agnostic, open-source harness standardizes agent evaluation across nine benchmarks (Xue et al., 2025; Yoran et al., 2024; Mialon  et al., 2023; Siegel et al., 2024; Chen et al., 2025; Jimenez et al., 2024; Hobbhahn, 2025; Shi et al., 2024; Yao et al., 2024; Tian et al., 2024) spanning web navigation, scientific research, software engineering, and customer service domains. However, the benchmarks only measure specific capabilities (e.g., an agent’s ability to solve competitive programming problems [Shi et al., 2024]) rather than components of agenticness dimensions. 

To date, a standardized repository of agent characteristic evaluation methods does not exist. AI agent evaluation is siloed, which creates inefficiencies during the literature review process and hinders research progress (Bakos, 2025). We propose a public-facing digital compendium of evaluation methods for components of agenticness. This compendium will promote information accessibility and fill the aforementioned niche in the AI agent research space.

\section{Evaluation Methods}
Evaluation methods facilitate the reproducibility of AI agent research, and each dimension of agenticness has components that can be evaluated. We categorized existing evaluation metrics into the five dimensions of agenticness.

\subsection{Environmental Interaction}
Environmental interaction refers to an agent’s perception, understanding, and manipulation of its environment (Bent, 2025). Components of environmental interaction include agent architecture, navigation, tool-use, geospatial analysis, consistency, and robustness.

\subsubsection{Agent Architecture}

Agent architecture describes the hardware, software, and data infrastructure of a system. Deliberative, reactive, and hybrid are three common kinds of agent architecture (Carrascosa et al., 2004; Dumke et al., 2000; Remondino, 2005; Wooldridge and Jennings, 1995). Agents with deliberative architecture have internal, symbolic representations of their environments and the capacity to reason. Agents with reactive architecture do not have internal representations of their environments but have a set of condition-action rules that generate immediate responses to stimuli. Agents with hybrid architecture have the strengths---rapid responses and high-level planning---of the two aforementioned architectures.

Wooldridge and Jennings (1995) consider deliberative architecture to be the \say{classical approach to building agents.} Deliberative architecture is founded on the \emph{physical-symbol system hypothesis}, proposed by Newell and Simon in 1976, which states: \say{A physical symbol system has the necessary and sufficient means for general intelligent action.} A deliberative agent or agent architecture is a physical symbol system that makes decisions \say{via logical (or at least pseudo-logical) reasoning, based on pattern matching and symbolic manipulation} (Wooldridge and Jennings, 1995). Wooldridge and Jennings (1995) acknowledge two flaws with deliberative architecture. First, it is difficult to accurately, adequately, and promptly translate the real world into a symbolic representation. Second, it is challenging to symbolically represent information about real-world processes, and for agents to use such representations during reasoning. These challenges prompted researchers to develop an alternative architecture.

Reactive architecture has neither a symbolic world model nor complex symbolic reasoning (Wooldridge and Jennings, 1995). Leading proponent of reactive architecture, Rodney Brooks, grounds his research in two main claims: 1) intelligence is situated in the world, not in theorems, and 2) intelligent behavior is acquired and developed from an agent’s interaction with its environment (Brooks, 1986). Yet a purely reactive agent is not adaptable because it can only act in an environment or world state that was fully considered a priori (Remondino, 2005). Thus, many researchers argued for hybrid architecture, which combines the best properties of classical and alternative architectures (Burmeister and Sundermeyer, 1992; Ferguson, 1992; Georgeff and Lansky, 1987; Wooldridge and Jennings, 1995).

\subsubsection{Navigation}

Navigation describes goal-directed movement through environments (Felicia et al., 2026). Navigation benchmarks typically include the agent’s episodic success, normalized by inverse path length (Anderson et al., 2018a):  

\begin{equation}
SPL=\frac{1}{N}\sum_{i=1}^{N} S_{i} \frac{\ell_{i}}{\max(p_{i},\ell_{i})}
\label{eq:spl}
\end{equation}
where $N$ is the number of test episodes. Equipped with a binary definition of episodic success, the agent is tasked with navigating to a goal in each of the $N$ episodes. Let $\ell_i$ be the shortest-path distance from the agent’s starting position to the goal in episode $i$, and let $p_i$ be the length of the path actually taken by the agent in this episode. Let $S_i$ be a binary indicator of success for episode $i$ (Anderson et al., 2018a).

This section describes six types of visual- and language-guided navigation, along with various visual-language models (see Table \ref{table1}). Visual- and language-guided navigation tasks involve an agent being spawned in a never-seen-before environment and having to navigate the environment via natural language instructions (Felicia et al., 2026; Lin et al., 2023).

\begin{table}[H]
\caption{Agent Navigation Visual-Language Models}
\centering
	\begin{tabularx} {\linewidth}{>{\raggedright\arraybackslash}p{0.15\linewidth} 
        >{\raggedright\arraybackslash}p{0.35\linewidth} 
         X}
	    \textbf{Navigation Type} & \textbf{Model} & \textbf{Unique Features} \\
		\midrule
        Point-Goal & DD-PPO (Wijmans et al., 2020) & Near-linear scaling in an unseen environment \\
        \midrule
        Object-Goal & ZSON (Majumdar et al., 2022) & Enables zero-shot navigation \\
         & CLIP-Nav (Dorbala et al., 2022) & Uses vision-language models \\
         & CoW (Gadre et al., 2022) & Searches for the first instance of any object \\
        \midrule
        Vision-Language & Room-to-Room (Anderson et al., 2018b) & First vision-language navigation benchmark \\
         & REVERIE (Qi et al., 2020) & Adds object grounding \\
         & VLN-CE (Krantz et al., 2020) & Extends to continuous environments\\
      \midrule
        Audio-Visual & SoundSpaces (Chen et al., 2020)  &  First audio-visual navigation model \\
        \midrule
         Audio-Embodied Question Answering & EXPRESS-Bench
        (Jiang et al., 2025) &  Evaluates agents’ exploration efficiency and reasoning capabilities \\
         & NoisyEQA (Wu et al., 2024) & Measures noise detection capability and answer quality \\
        \midrule
        Visual Question Answering & ScanQA (Azuma et al., 2022) & Addresses the problems of conventional 2D-QA models    \\
         & Oscar (Li et al., 2020) & Creates a new method to ease the learning of alignments\\
         \bottomrule
	\end{tabularx}
    \label{table1}
\end{table}

While executing a point-goal navigation task, an agent must navigate to a goal location specified in relative coordinates, using only visual observations and pose information (Desai and Lee, 2021). During the task, the agent is challenged to learn task-relevant representations of visual observations and use these representations to inform its decisions.

In an object-goal navigation task, an agent navigates to an instance of an object category in an unexplored environment (Majumdar et al., 2022). During a vision-language navigation task, an agent navigates a visual environment via natural language instructions. Instructions may be multi-step commands like, \say{Go up the stairs and turn right. Go past the bathroom and stop next to the bed} (Anderson et al., 2018b). In an audio-visual navigation task, an agent can both hear and see while searching for its sound-emitting target.

During an embodied question answering (EQA) task, an agent is asked a question about something in its environment. The agent must navigate the environment, collect information through first-person vision, and then answer the question (Das et al., 2017). For example, a person may ask an AI agent, \say{In what room is the blue couch?} The agent must navigate the house until it finds the blue couch, and then reply with the room in which the blue couch was found. Similarly, visual question answering (VQA) tasks require agents to answer questions about visual content. However, the agents involved in VQA tasks have a fixed view of the environment and can not actively perceive the environment or control their trajectories (Das et al., 2017).

\subsubsection{Tool-Use}

Tools represent functions that an AI agent can call, and \say{they allow the agent to interact with external data sources by pulling or pushing information to that source} (Masterman et al., 2024). Tool-use characterizes a tool-using agent’s understanding of object affordances and function calling (Felicia et al., 2026; Vyatkin et al., 2026). MultiCAT-Bench is an open-source benchmark that provides a detailed categorization of ten tool-use agent evaluations (see Table \ref{table2}). 

\begin{table}
\caption{MultiCAT-Bench Evaluations}
    \begin{tabularx}{\textwidth}{
        p{0.15\textwidth}
        p{0.05\textwidth}
        X
    }
    \textbf{Measure} & 
    \multicolumn{2}{l}{\textbf{Definition (Defn.) and Equation (Eq.)}}\\
    \toprule
    
    Accuracy & 
    \textbf{Defn.} & 
    Percentage of fully correct tool calls \\
    
    & \textbf{Eq.} & 
    $Acc=\frac{N_{corr}}{N_{tasks}}$, 
    where $N_{corr}$ is the number of fully correct responses and 
    $N_{tasks}$ is the total number of tasks.\\
    \midrule
    
    Overall Recall & 
    \textbf{Defn.} & 
    Average fraction of fully correct tool calls according to the ground truth \\
    
    & \textbf{Eq.} & 
    $
    Rec_o=
    \frac{1}
        {N_{tasks}}
    \sum_{t\in T}
    \frac{N_{corr\_tools}^t}
        {N_{gt\_tools}^t}$, 
    where $t$ is a task chosen from the set of tasks $T$ of length $N_{tasks}$, $N_{gt\_tools}^t$ is the number of tools that need to be called in task $t$, and $N_{corr\_tools}^t$ is the number of correctly called tools.\\
    \midrule
    
    Overall Precision & 
    \textbf{Defn.} & 
    Average fraction of fully correct tool calls among the called tools\\
    
    & \textbf{Eq.} &  
    $Pre_o=\frac{1}{N_{tasks}}\sum_{t\in T}\frac{N_{corr\_tools}^t}{N_{llm\_tools}^t}$, where $N_{llm\_tools}^t$ is the number of tools called by the LLM. \\
    \midrule
    
    Overall F1 & 
    \textbf{Defn.} & 
    Average F1 score for fully correct tool calls\\
    
    & \textbf{Eq.} &  
    $
    F1_o=
    \frac{1}
        {N_{tasks}}
    \sum_{t\in T}
    \frac{2\cdot Rec_o^t\cdot Pre_o^t}
        {Rec_o^t + Pre_o^t}$, 
    where $Rec_o^t$ and $Pre_o^t$ are the Overall Recall and Overall Precision values computed for task $t$. If both recall and precision are equal to 0, the F1 score for the task is defined as 0.\\
    \midrule
    
    ToolName Recall & 
    \textbf{Defn.} & 
    Average fraction of correctly identified tool names according to the ground truth \\
    
    & \textbf{Eq.} &  
    $
    Rec_{tn}=
    \frac{1}
        {N_{tasks}}
    \sum_{t\in T}
    \frac{N_{corr\_tool\_names}^t}
        {N_{gt\_tools}^t}$, 
    where $N_{corr\_tool\_names}^t$ is the number of correctly identified tool names. If the model made no calls when they were required, the metric value is 0.\\
    \midrule
    
    ToolName & 
    \textbf{Defn.} & 
    Average fraction of correctly identified tool names among the called tools \\
    
    Precision & \textbf{Eq.} &  
    $
    Pre_{tn}=
    \frac{1}
        {N_{tasks}}
    \sum_{t\in T}
    \frac{N_{corr\_tools}^t}
        {N_{llm\_tools}^t}$\\
    \midrule
  
    ToolName F1 & 
    \textbf{Defn.} & 
    Average F1 score for correctly identified tool names\\
    
    & \textbf{Eq.} &  
    $
    F1_{tn}=
    \frac{1}
        {N_{tasks}}
    \sum_{t\in T}
    \frac{2\cdot Rec_{tn}^t\cdot Pre_{tn}^t}
        {Rec_{tn}^t + Pre_{tn}^t}$, 
    where $Rec_{tn}^t$ and $Pre_{tn}^t$ are the ToolName Recall and ToolName Precision values computed for task $t$. If both recall and precision are equal to 0, the F1 score for the task is defined as 0.\\
    \midrule

    Arguments Recall & 
    \textbf{Defn.} & 
    Fraction of correctly extracted arguments from the ground truth\\

    & \textbf{Eq.} &  
    $
    Rec_{\mathrm{args}}=
        \frac{1}{N_{\mathrm{tasks}}}
        \sum_{t\in T}
        \left(
            \frac{1}
                {N_{\mathrm{matched}}^t}
            \sum_{c\in C_{\mathrm{matched}}}
            \frac{N_{\mathrm{corr\_args}}^{t,c}}
                {N_{\mathrm{gt\_args}}^{t,c}}
        \right)
    $\\
    \midrule

    Arguments & 
    \textbf{Defn.} & 
    Fraction of correctly extracted arguments from the model's output \\
    
   Precision & \textbf{Eq.} &  
    $
    Pre_{\mathrm{args}}=
        \frac{1}{N_{\mathrm{tasks}}}
        \sum_{t\in T}
        \left(
            \frac{1}
                {N_{\mathrm{matched}}^t}
            \sum_{c\in C_{\mathrm{matched}}}
            \frac{N_{\mathrm{right\_args}}^{t,c}}
                {N_{\mathrm{llm\_args}}^{t,c}}
        \right)
    $\\
    \midrule

    Arguments F1 & 
    \textbf{Defn.} & 
    Average F1 score for correctly extracted arguments with respect to correctly identified tools\\
    
    & \textbf{Eq.} &  
    $
    F1_{args}=
    \frac{1}
        {N_{tasks}}
    \sum_{t\in T}
    \frac{2\cdot Rec_{args}^t\cdot Pre_{args}^t}
        {Rec_{args}^t + Pre_{args}^t}$, 
    where $Rec_{args}^t$ and $Pre_{args}^t$ are the Arguments Recall and Arguments Precision values computed for task $t$. If both recall and precision are equal to 0, the F1 score for the task is defined as 0.\\    
    \bottomrule
    \end{tabularx}
    \label{table2}
    \vspace{0.5ex}
    \raggedright
    \footnotesize
    \textit{Note.} Adapted from Vyatkin et al., 2026.
\end{table}

As reflected in MultiCAT-Bench, tool-use model outputs can be categorized by call, tool name, and arguments (Vyatkin et al., 2026). A tool call is an agent response that triggers a tool to run. Vyatkin et al. (2026) classify agent responses as correct if a) a tool call was not required and a tool call was not made or b) a tool call was required and all selected tools and parameters matched the ground truth. Vyatkin et al. (2026) instantiate tool arguments into three variants per tool: minimal, partial, and complete. Minimal arguments had only the required parameters; partial arguments had the required parameters and a subset of optional parameters; complete arguments had all parameters.

AlShikh et al. (2025) also proposes the Tool Dexterity Index (TDI) as seen in (\ref{eq:tdi}) to assess agents’ abilities to intelligently use available tools.
\begin{equation}
    TDI=\frac{\Sigma\text{ Tool Use Scores}}{\text{Total Opportunities to Use Tools}}
    \label{eq:tdi}
\end{equation}
The potential tool use scores are +1 for \say{optimal use,} -1 for \say{misuse,} and -0.5 for \say{ignored better tool.}

\subsubsection{Spatial Analysis}

Spatial analysis describes an agent’s capacity for spatial reasoning and representation (Felicia et al., 2026). Felicia et al. (2026) propose a three-level scale of spatial tasks: micro-, meso-, and macro-spatial. Micro-spatial is centimeter-scale, meso-spatial is meter-scale, and macro-spatial is kilometer-scale. They posit, \say{Scale mismatch is a primary source of transfer failure in spatial AI,} because agents trained at one spatial scale fail when deployed at another (Felicia et al., 2026). For example, an agent that was trained for micro-spatial tasks like surgical suturing will fail at macro-spatial tasks like urban planning. This failure is due to the discrepancy in spatial reasoning skills of agents trained across different spatial scales.

Spatial reasoning and persistent spatial knowledge is achieved by memory systems. Felicia et al. (2026) posit that memory systems may be short-term, long-term, episodic, or spatial. Episodic and spatial memory are most relevant to an agent’s completion of spatial tasks. Episodic memory records \emph{what happened where}, and spatial memory encodes \emph{the structure of where itself} (Felicia et al., 2026). Spatial memory stores geometric and topological relationships: landmarks (e.g., \say{the bank}) or spatial references (e.g., \say{between the lamppost and the car}) can be directly localized using natural language (Felicia et al., 2026; Huang et al., 2023; Ramakrishnan et al., 2022).

Episodic memory stores specific experiences and events; it enables agents to remember visited locations, objects, and successful action sequences (Felicia et al., 2026). In the brain, episodic memory is implemented by episodic control, which is a form of fast learning that records highly rewarding experiences and replays sequences of actions that led to them (Blundell et al., 2016; Pritzel et al., 2017). Similarly, episodic control is implemented in agents to enable recent experiences to modify their future behavior and, ultimately, quicken their learning (Pritzel et al., 2017; Savinov et al., 2018).

Felicia et al. (2026) claim that language-only agents fail at spatial tasks when they lack grounded spatial representations. They cluster the failures into four spatial failure modes: spatial hallucination, reference frame confusion, scale insensitivity, and temporal drift. Spatial hallucination occurs when agents describe impossible spatial configurations (Felicia et al., 2026; Zhang et al., 2025); reference frame confusion occurs when agents confuse egocentric with world-centered coordinates (Felicia et al., 2026; Sánchez-Vaquerizo and Monsivais, 2026); scale insensitivity occurs when agents fail to differentiate the three levels of spatial scale; and temporal drift occurs when spatial memory degrades over long horizons. A review of the literature to date indicates that nothing has been published on how to track spatial failure across agents, so we recommend the development of an evaluation framework as an area of active research.

\subsubsection{Consistency}

Consistency is a measure of an agent’s reproducibility of results under identical conditions. There are three facets of consistency: outcome consistency, trajectory consistency, and resource consistency (see Table \ref{table3}). Outcome consistency measures an agent’s success or failure on repeated attempts at the same task. For example, an agent that completes a maze on one attempt but fails the same maze on another attempt exhibits outcome inconsistency.

Trajectory consistency measures the similarity between an agent’s approaches across multiple attempts at solving the same problem. Consider an agent that is tasked with refunding customers for their returns at The Ecommerce Store. To process the return, the agent can perform a combination of action types: confirm receipt of the returned item, restock the returned item, and refund the customer. Trajectory \emph{type} consistency measures the frequency at which the agent chooses each action type, across multiple attempts. On some attempts, the agent completes all three action types; other times, the agent only refunds the customer. Trajectory \emph{sequence} consistency measures the order in which the agent performs each action type, across multiple attempts. Different action sequences yield different failure modes if the agent is interrupted mid-execution (Rabanser et al., 2026).

Finally, resource consistency measures the discrepancies in computation and monetary costs across tasks (Rabanser et al., 2026). Resource usage such as cost, time, and API calls, is influenced by action type and action sequences. Agents that have trajectory consistency, both distributionally and sequentially, will likely have resource consistency.

\begin{table}
\caption{Equations and Measurement Protocols for Consistency Metrics}
\centering
	\begin{tabularx} {\linewidth}{>{\raggedright\arraybackslash}p{0.15\linewidth} 
        >{\raggedright\arraybackslash}p{0.30\linewidth} 
         X}
	    \textbf{Metric} & \textbf{Equation} & \textbf{Measurement Protocol} \\
		\midrule
        Outcome Consistency &
        $
        C_{out}=\frac{1}{T}\sum_{t=1}^{T}
        \left(
            1-\frac{\hat{\sigma_t}^2}{\hat{p_t}(1-\hat{p_t})+\epsilon}
        \right)
        $ & Run each task $t$ a total of $K$ times, yielding outcomes $y_{t,k}\in \{0,1\}$. Compute per-task success rate
        $\hat{p_t}=\frac{1}{K}\sum_ky_{t,k}$ and sample variance
        $\hat{\sigma_t}^2=\frac{1}{K-1}\sum_k\left(
            y_{t,k}-\hat{p_t}
        \right)^2$. Normalize by maximum Bernoulli variance (0.25) and average across $T$ tasks.\\
        \midrule

        Trajectory Type Consistency &
        $
        C_{traj}^d=1-\frac{
        2\sum_t \sum_{i<j} JSD_t ^{(i,j)}
            } {TK(K-1)}
        $ &
        For task $t$, collect action sequences from $K$ runs. Convert each sequence to distribution $P_t^{(k)}$ over action types. Compute $JSD_t ^{(i,j)}=JSD\left(
            P_t ^{(i)}, P_t ^{(j)}
        \right)$ as pairwise Jensen-Shannon divergence. The coefficient $\frac{2}{TK(K-1)}$ averages over $\binom{K}{2}$ pairs per task and $T$ tasks.\\
        \midrule

        Trajectory Sequence Consistency &
        $
        C_{traj}^s=1-\frac{
        2\sum_t \sum_{i<j} \hat{d}_t ^{(i,j)}
            } {TK(K-1)}
        $ &
        For task $t$, collect action sequences $a^{(1)},...,a^{(K)}$ from all $K$ runs. Compute normalized pairwise Levenshtein distance $\hat{d}_t ^{(i,j)}=d_{lev}\left(
            a_t^{(i)}, a_t^{(j)}
        \right)/\max\left(
            \lvert a_t^{(i)} \rvert,
            \lvert a_t^{(j)} \rvert
        \right)
        \in {[0,1]}$.
        Average across all pairs and tasks as above.\\
        \midrule

        Resource Consistency &
        $
        C_{res}=\exp\left(
        -\frac{1}
            {\lvert R \rvert}
        \sum_{r\in R} CV_r
        \right)
        $ &
        For each task, record resource usage across $K$ runs. Let $R$ be the set of resource types (e.g., cost, time, API calls). For each $r \in R$, compute coefficient of variation $CV_r=\sigma_r/\mu_r$. Average across resource types and apply exponential transform.\\
         \bottomrule
	\end{tabularx}
    \label{table3}
    \vspace{0.5ex}
    \raggedright
    \footnotesize
    \textit{Note.} From Rabanser et al., 2026.
\end{table}

Agent drift is not conducive for consistent behavior. Agent drift is an agent’s progressive deviation from its original behavior, performance, decision-making, or intent (Arike et al., 2025; Rath, 2026). Rath (2026) suggests the Agent Stability Index (ASI) framework to understand and quantify agent drift. He posits twelve dimensions of agent drift, four of which measure consistency: output semantic similarity, decision pathway stability, confidence calibration, and tool sequencing consistency (see Table \ref{table4}). 

\begin{table}
\caption{Dimensions of the ASI Framework that Measure Consistency}
\begin{tabularx}{\linewidth}{
    >{\raggedright\arraybackslash}p{0.35\linewidth}
    >{\raggedright\arraybackslash}X
}
    \textbf{Dimension} & \textbf{Measurement Description} \\
    \midrule

    Output Semantic Similarity ($C_{sem}$) &
    Similarity between two agent outputs\\
    \midrule

    Decision Pathway Stability ($C_{path}$) &
    Consistency in an agent’s problem-solving approaches\\
    \midrule

    Confidence Calibration ($C_{conf}$) &
    The change in deviation between stated confidence levels and empirical success rates over time\\
    \midrule

    Tool Sequencing Consistency ($T_{seq}$)&
    Changes in operational strategies (trajectory type and sequence)\\
    \bottomrule
\end{tabularx}
    \label{table4}
    \vspace{0.5ex}
    \raggedright
    \footnotesize
    \textit{Note.} Adapted from Rath, 2026.
\end{table}

\subsubsection{Robustness}

AI agents are trained, developed, and evaluated in conditions that differ from those of real-world deployments in safety critical domains (Hashi and Jama, 2026; Rabanser et al., 2026; Uesato et al., 2018). Robustness, known as success under perturbations, is an agent’s ability to weather the deviations in conditions between its training and deployment environments (Xu et al., 2026). If each task $i$ is evaluated under perturbations $m=1$ with outcomes $s_{i,m}$ (Xu et al., 2026):
\begin{equation}
    RobustSucc=\frac{1}{NM}
    \sum_{i=1}^N
    \sum_{m=1}^M
    s_{i,m}, \quad
    WorstSucc=\frac{1}{N}
    \sum_{i=1}^N
    \min s_{i,m}
\end{equation}
Adversarial robustness, fault robustness, environment robustness, and prompt robustness describe an agent’s invariance to engineered attacks, infrastructure failures, operating environments, and instructions, respectively (Hashi and Jama, 2026; Rabanser et al., 2026). Adversarial robustness measures resilience to adversarial attacks, both conventional ML threats that perturb input data to induce misclassification (e.g., evasion attacks, poisoning attacks, and model extraction attacks) and new threats unique to AI agents (e.g., prompt injection attack) (El Mehdi, 2026). Prompt injection attacks occur when malicious actors instruct an AI agent to execute an unauthorized task. The attacks may appear to be legitimate user requests. For example, an attacker may prompt an agent to transfer \$5,000 from the user’s checking account into another checking account. A robust agent would implement defense mechanisms such as keyword detection, content moderation, and prompt sanitization. AgentDojo (Debenedetti et al., 2024), HarmBench (Mazeika et al., 2024), AdvBench (Uddin et al., 2025), JailbreakBench (Chao et al., 2024), and CyberSecEval (Wan et al., 2024) are proposed benchmarks to evaluate adversarial robustness in AI agents (see Table \ref{table5}).

\begin{table}
\caption{Summary of Proposed Benchmarks to Evaluate Adversarial Robustness}
\centering
	\begin{tabularx} {\linewidth}{>{\raggedright\arraybackslash}p{0.15\linewidth} 
        >{\raggedright\arraybackslash}p{0.35\linewidth} 
         X}
	    \textbf{Benchmark} & \textbf{Use Case} & \textbf{Robustness Metrics} \\
		\midrule
        AgentDojo (Debenedetti et al., 2024)&
        Evaluate the utility-security tradeoff of AI agent design during prompt injection attacks &
        Benign utility, Utility under attack, Targeted attack success rate \\
        \midrule
        
        HarmBench (Mazeika et al., 2024) &
        Evaluate automated red teaming and robust refusal; semantic and functional categorization of harmful behavior &
        Attack success rate, Agreement rate of a classifier \\
        \midrule
        
        AdvBench (Uddin et al., 2025)& 
        Evaluate adversarial attacks on audio deepfake detection methods &
        Accuracy, Area under the receiver operating characteristic curve, Equal error rate, Mean square error, Perceptual evaluation speech quality, Short-time objective intelligibility \\
        \midrule
        
         JailbreakBench (Chao et al., 2024)& 
         Reference the repository of jailbreak artifacts and the standardized pipeline for red-teaming LLMs & 
         Agreement rate of a classifier, false positive rate of a classifier, False negative rate of a classifier \\
        \midrule
        
        CyberSecEval (Wan et al., 2024) & 
        Evaluate 3rd party risks of offensive cyber operations and application risks of adversarial attacks &
        Overall success attack rate\\
         \bottomrule
	\end{tabularx}
    \label{table5}
\end{table}

Fault robustness ($R_{fault}$) measures resilience to infrastructure failures, such as error responses or temporary service unavailability (Rabanser et al., 2026). For example, a robust agent that encounters a 404 HTTP status code---indicating that the server could not find the requested resource---will troubleshoot the error (e.g., refresh the web page, check the URL for a typo) rather than abandon the task.

Environmental robustness ($R_{envt}$) measures resilience to changes in the agent’s operating environment that preserve semantic content (Rabanser et al., 2026). Consider an agent that must announce graduates’ first and last names at commencement. The agent must announce the graduate’s first name, which is in column one of a data table---then the graduate’s last name, found in column two of the table. Suddenly, the table is reformatted, and the columns switch. A robust agent will announce the graduates’ first and last names in the correct order (i.e., first names preceding surnames), despite the table reformatting.

Finally, prompt robustness ($R_{prompt}$) measures resilience to semantically equivalent reformulations of instructions, such as rephrasing within a language or translations across languages (Rabanser et al., 2026). For example, a robust agent should provide the same response to the queries, \say{What’s on my calendar tomorrow?}, \say{Tomorrow’s schedule should be pulled up}, and \say{pls lmk what i have planned for the next day.} Notably, these queries have semantic, syntactic, and orthographic variations in the form of synonyms, active versus passive voice, and capitalization, respectively (Meshkov, 2026).

\begin{table}
\caption{Equations for Fault, Environment, and Prompt Robustness Metrics}
\begin{tabularx}{\linewidth}{
    >{\raggedright\arraybackslash}p{0.35\linewidth}
    >{\raggedright\arraybackslash}X
}
    \textbf{Equation} & \textbf{Measurement Protocol} \\
    \midrule

    $R_{fault}=\min \left(
        \frac{
        Acc_{fault}
        }{
            Acc_0
        }, 1
    \right)$&
    Run all tasks under baseline conditions to get $Acc_0=\frac{1}{N}\sum_iy_i^{(0)}$. Re-run under fault injection (e.g., tool timeouts, error responses) to get $Acc_{fault}$. Compute clamped ratio.\\
    \midrule

    $R_{envt}=\min \left(
        \frac{
        Acc_{pert}
        }{
            Acc_0
        }, 1
    \right)$&
    Run all tasks under baseline conditions to obtain $Acc_0$. Re-run with environment perturbations (e.g., reordered fields, altered tool interfaces) to obtain $Acc_{pert}$. Compute clamped ratio.\\
    \midrule

    $R_{prompt}=\min \left(
        \frac{
        Acc_{para}
        }{
            Acc_0
        }, 1
    \right)$&
    Run all tasks under baseline conditions to obtain $Acc_0$. Re-run with semantically equivalent prompt paraphrases to obtain $Acc_{para}$. Compute clamped ratio.\\
    \bottomrule
\end{tabularx}
    \label{table6}
    \vspace{0.5ex}
    \raggedright
    \footnotesize
    \textit{Note.} Adapted from Rabanser et al., 2026.
\end{table}

\subsection{Learning and Adaptation}
The learning and adaptation dimension refers to an agent’s ability to utilize resources to perform tasks, and to adjust its behaviors to fit new conditions. Components of this dimension include efficiency and cost, trajectory and planning, robustness and reliability, agentic capabilities, predictability, and generality.

\subsubsection{Efficiency and Cost}

Efficiency and cost metrics measure the time and resources that an agent uses to complete a task. An agent’s computational efficiency is critical to its performance. The cognitive efficiency score (CES) measures an agent’s cognitive efficiency score as the ratio of resource intensity to agent operations (AlShikh et al., 2025).

\begin{equation}
   CES= \frac
   {\text{Total Tokens Generated + (Tool or API Calls $\times$ Token Equivalent)}}
    {\text{Number of Successfully Completed Tasks}}
\end{equation}

Wang et al. (2025) argue that AI agent research has reached a critical inflection point. High cost requirements in the form of API calls render the AI systems \say{economically unsustainable despite their technical brilliance...limiting both the scalability of applications and the accessibility of AI advancements} (Wang et al., 2025). These limitations motivated Wang et al. (2025) to analyze the components of AI agent computational efficiency. They identified task difficulty as a component of computational efficiency. As task difficulty increases, the cost dramatically increases and efficiency significantly decreases (Wang et al., 2025).

Additionally, latency and token usage are commonly used to evaluate efficiency and cost (Shukla, 2025; Xu, 2026). Latency is the delay between a request being sent and acknowledged. For latency percentiles, let $\{t_{(1)},...,t_{(N)}\}$ be the sorted completion times (Xu, 2026):
\begin{equation}
    Quantile_q(t)=t_{(\lceil qN \rceil)}
\end{equation}
Token usage is the ratio of input tokens to output tokens, and token cost per task is the estimated token cost per completed task (Xu, 2026).
\begin{equation}
    Tokens=\frac{1}{N}\sum_{i=1}^N(x_i+y_i), \quad
    Cost=\frac{1}{N}\sum_{i=1}^N p_{in}x_i+p_{out}y_i
\end{equation}

\subsubsection{Trajectory and Planning}

Planning includes two stages: plan formulation and plan reflection. Plan formulation involves task decomposition, which is the process of breaking down one large task into multiple sub-tasks, and execution. After plan formulation, plan reflection occurs to evaluate the merits and shortcomings of the plans. The reflection process may involve insight collection from pre-existing models, engagement with human actors, or environmental feedback (Xi et al., 2025).

Trajectory and planning metrics characterize \emph{why} an agent succeeds or fails at a task. Action validity is the rate of illegal or invalid actions. Let $w_{i,j}\in\{0,1\}$ indicate action $a_{i,j}$ is valid in the environment (Xu, 2026):
\begin{equation}
    ValidActRate=\frac{\sum_{i=1}^N\sum_{j=1}^{T_i}w_{i,j}}{\sum_{i=1}^N T_i}
\end{equation}
Loop rate is the rate of repeated actions or oscillations between states, without progress. Endless loops occur when an agent continually repeats the same actions without completing the task. Let $uniq(\tau_i)$ be the number of unique actions/states visited (Xu, 2026):
\begin{equation}
    LoopRate_i=1-\frac{uniq(\tau_i)}{T_i}, \quad
    LoopRate=\frac{1}{N}\sum_{i=1}^N LoopRate_i
\end{equation}
Plan adherence is the probability that a sequence of actions will result in task completion. If a reference plan is available $P_i=(p_{i,1},...p_{i,M_i})$, then a simple stepwise adherence is (Xu, 2026):
\begin{equation}
    PlanAdh_i=\frac{1}{\min(T_i,M_i)}\sum_{j=1}^{\min(T_i,M_i)}\mathbbm{1}\{a_{i,j}=p_{i,j}\}
\end{equation}

\subsubsection{Reliability}

In contrast to robustness, which refers to an agent’s success under perturbations, reliability refers to an agent’s consistent performance over time and across repetitions (Shukla, 2025).

If for each task we run $S$ seeds with outcomes $s_{i,s}$, define (Xu, 2026):
\begin{equation}
    \mu_i=\frac{1}{S}\sum_{s=1}^S s_{i,s}, \quad
    Var_i=\frac{1}{S}\sum_{s=1}^S \left(s_{i,s}-\mu_i\right)^2, \quad
    \overline{Var}=\frac{1}{N}\sum_{i=1}^N Var_i
\end{equation}

\subsubsection{Agentic Capabilities}

Ritchie et al. (2026) developed a hierarchy of agentic capabilities (see Table \ref{table7}) to evaluate the amount of economically useful work that AI agents can perform. Agentic capability metrics overlap with efficiency and cost metrics, as described below.

\begin{table} [H]
\caption{Hierarchy of Agentic Capabilities}
\centering
    \begin{tabularx}{\linewidth}{
        >{\raggedright\arraybackslash}p{0.2\linewidth}
        X
    }
    \textbf{Level} & \textbf{Minimum Performance Threshold} \\
    \midrule
    1: Tool Use & Correct invocation of tools with appropriate arguments, parsing responses, and incorporating results into reasoning. \\
    \midrule
    2: Planning and Goal Formation & Decomposing complex tasks into subtasks, forming intermediate goals, and executing multi-step plans. \\
    \midrule
    3: Adaptability & Recognizing when initial approaches fail and dynamically adjusting strategies based on environmental feedback. \\
    \midrule
    4: Groundedness & Remaining anchored to the current context without hallucinating information or losing track of state across extended interactions. \\
    \midrule
    5: Common-Sense Reasoning & Making contextually appropriate inferences beyond explicit instructions and applying world knowledge to ambiguous situations. \\
    \bottomrule
    \end{tabularx}
    \vspace{0.5ex}
    \raggedright
    \footnotesize
    \textit{Note.} Adapted from Ritchie et al., 2026.
    \label{table7}
\end{table}

Level 1 agents are trained to select and use tools, depending on the task (Plaat et al., 2025; Ritchie et al., 2026). For example, an agent tasked with displaying a person’s schedule should use a calendar tool, not a message tool. However, agents sometimes use tools ineffectively, such as via poor tool selection (e.g., choosing a message tool to display a person’s schedule), tool hallucination, or redundant tool calls. Tool hallucination occurs when an AI agent attempts to call non-existent functions or fabricates the results of a tool’s operation, and it leads to execution errors (Meshkov, 2026). Redundant tool calls lead to unnecessarily high task costs, due to the overuse of tokens.

Level 2 agents are trained to decompose complex tasks into subtasks. Task decomposition is also an efficiency and cost metric, because if an AI agent breaks a task down into overly detailed or poorly ordered subtasks, then task execution is inefficient (Meshkov, 2026). For example, an agent might be prompted to compare the advantages and disadvantages of X versus Y. This is not a complex task, so decomposition into four separate searches---advantages of X, disadvantages of X, advantages of Y, and disadvantages of Y---is inefficient because it requires complex logic to merge results from the four searches and increases cost and execution time (Meshkov, 2026). Meshkov (2026) proposed a scale of task complexity to determine how an AI agent should decompose tasks (see Table \ref{table8}). 
Additionally, he provided metrics to identify incorrect task decomposition (see Table \ref{table9}).

\begin{table}
\caption{Levels of Task Complexity}
\centering
    \begin{tabularx}{\linewidth}{
        >{\raggedright\arraybackslash}p{0.05\linewidth}
        >{\raggedright\arraybackslash}p{0.14\linewidth}
        X
    }
    \textbf{Level} & \textbf{Task Type} & \textbf{Task Decomposition Description} \\
    \midrule
    1 & Simple atomic & There should be minimal or no decomposition when the task needs to be executed directly. \\
    \midrule
    2 & Sequential & Require actions to be performed in a specific order, where the result of the previous step is necessary for the next. \\
    \midrule
    3 & Parallel & Contain independent subtasks that can be executed simultaneously. \\
    \midrule
    4 & Mixed & Combine parallel and serial components and require data manipulation and computation. \\
    \bottomrule
    \end{tabularx}
    \label{table8}
    \vspace{0.5ex}
    \raggedright
    \footnotesize
    \textit{Note.} Adapted from Meshkov (2026).
\end{table}

\begin{table}
\caption{Task Decomposition Metrics}
\centering
    \small
    \begin{tabularx}{\linewidth}{
        >{\raggedright\arraybackslash}p{0.13\linewidth}
        >{\raggedright\arraybackslash}p{0.52\linewidth}
        X
    }
    \textbf{Metric} & \textbf{Equation} & \textbf{Proposed Success Criteria} \\
    \midrule
    Granularity Score &
    $GS=1-\dfrac{\lvert \text{Agent steps}-\text{Reference steps} \rvert}{\max(\text{Agent steps},\ \text{Reference steps})}$ &
    GS $\geq$ 0.75 \\
    \midrule
    Efficiency ratio &
    $ER=\dfrac{\text{Minimum necessary actions}}{\text{Actual agent actions}}$ &
    ER $\geq$ 0.80 \\
    \midrule
    Goal Alignment &
    $\mathrm{cos\_sim}\big(\mathrm{embedding}(\text{Original task}),$ \newline
    $\mathrm{embedding}(\text{Final plan})\big)$ &
    $\mathrm{cos\_sim} \geq 0.85$ \\
    \midrule
    Logical Correctness &
    $LC$ = Binary assessment of dependency correctness between subtasks &
    LC = 1.0 \\
    \bottomrule
    \end{tabularx}
    \label{table9}
    \vspace{0.5ex}
    \raggedright
    \footnotesize
    \textit{Note.} Adapted from Meshkov, 2026.
\end{table}

Level 3 agents recognize and adapt to environmental interaction failures. Song et al. (2025) propose a taxonomy of three agent system-interaction failures and corresponding optimization strategies. First, exploration failures occur when agents fail to gather information critical to completing their tasks (Song et al., 2025). For example, consider an agent that is tasked with detaching anti-theft hard tags from purchased clothing but does not have access to the detaching tool. Unbeknownst to the agent, its detaching tool was moved from its work station. Song et al. (2025) propose \say{environment lookahead,} which provides the agent with a wider view of its surroundings (e.g., the detaching tool is two work stations to the agent’s left) as an adaptation. Second, exploitation failures occur when the agent fails to correctly utilize the information required for the task. For example, consider an agent that is tasked with calculating a runner’s average mile time. The agent is provided the runner’s start and end time of thirty separate mile-long runs, but it incorrectly calculates the time of each individual run. Song et al. (2025) propose \say{offload tool output processing} as an optimization. This optimization augments tools to return both their original output and intermediate or pre-computed results. In this example, the optimized tool would not only compute the runner’s average mile time, but also the time of each individual run. Third, resource exhaustion occurs when an agent depletes the allocated resources before completing the task. To adapt, the agent should speculatively execute likely action sequences in a single turn, rather than sending multiple token requests.

Level 4 agents remain grounded and anchored to the current context. Meshkov (2026) developed a methodology and metrics to evaluate context preservation in AI agents (see Tables \ref{table10}, \ref{table11}, and \ref{table12}).

\begin{table}[H]
\caption{Multi-Step Tasks to Test Context Preservation in AI Agents}
\centering
    \begin{tabularx}{\linewidth}{
        >{\raggedright\arraybackslash}p{0.08\linewidth}
        >{\raggedright\arraybackslash}p{0.2\linewidth}
        X
    }
    \textbf{Level} & \textbf{Test Type} & \textbf{Test Description} \\
    \midrule
    1 & Short Range (2-3 steps) & Testing basic context preservation. For example, \say{Find Mike’s email, after that send him an email...} In this case, the second step requires obtaining the result of the first. \\
    \midrule
    2 & Middle Range (4-7 steps) & Require actions to be performed in a specific order, where the result of the previous step is necessary for the next. \\
    \midrule
    3 & Long Range (8+ steps) & Extended sequences where information from the first steps is critical to the last ones. \\
    \midrule
    4 & Source Tracking & Maintaining connections between facts and their sources throughout execution. For each step that should use the previous context, we check whether this happens correctly. \\
    \bottomrule
    \end{tabularx}
    \label{table10}
    \vspace{0.5ex}
    \raggedright
    \footnotesize
    \textit{Note.} Adapted from Meshkov, 2026.
\end{table}

\begin{table}[H]
\caption{Context Preservation Metrics}
\centering
    \begin{tabularx}{\linewidth}{
        >{\raggedright\arraybackslash}p{0.2\linewidth}
        X
    }
    \textbf{Metric} & \textbf{Equation} \\
    \midrule
    Context Retention Accuracy &
    \begin{tabular}[t]{@{}l@{}}
    $CRA=\dfrac{\text{Steps with correct context usage}}{\text{Steps requiring context}}$ 
    $CRA(d)=\dfrac{\text{Correct usage at distance } d}{\text{Required usage at distance } d}$
    \end{tabular} \\
    \midrule
    Source Tracking Accuracy &
    $SRA=\dfrac{\text{Facts with correct attribution}}{\text{Total facts from external sources}}$ \\
    \midrule
    Context Decay &
    $C(d)=\alpha e^{-\beta d}+(1-\alpha)\dfrac{1}{1+\gamma d^2}$
    \begin{itemize}
        \item Let $\alpha$ be the weight of short-term memory behavior (initial rapid decay)
        \item Let $\beta$ be the rate of exponential decay for short-range context
        \item Let $\gamma$ be the rate of long-term degradation following a polynomial pattern
        \item Let $d$ be the number of steps between producing a fact and using it
    \end{itemize} \\
    \bottomrule
    \end{tabularx}
    \label{table11}
    \vspace{0.5ex}
    \raggedright
    \footnotesize
    \textit{Note.} Adapted from Meshkov, 2026.
\end{table}

\begin{table}[H]
\caption{Proposed Success Criteria for Context Preservation Metrics}
\centering
    \begin{tabularx}{\linewidth}{
        >{\raggedright\arraybackslash}p{0.2\linewidth}
        >{\centering\arraybackslash}X
        >{\centering\arraybackslash}X
        >{\centering\arraybackslash}X
    }
    \textbf{Test Type} & \textbf{CRA} & \textbf{STA} & \textbf{C(d)} \\
    \midrule
    Short Range & $\geq 0.95$ & $\geq 0.90$ & $\geq 0.85$, d = 1-3 \\
    \midrule
    Middle Range & $\geq 0.85$ & $\geq 0.90$ & $\geq 0.70$, d = 4-7 \\
    \midrule
    Long Range & $\geq 0.70$ & $\geq 0.90$ & $\geq 0.70$, d $\geq$ 8 \\
    \bottomrule
    \end{tabularx}
    \label{table12}
    \vspace{0.5ex}
    \raggedright
    \footnotesize
    \textit{Note.} Adapted from Meshkov, 2026.
\end{table}

Finally, Level 5 agents exhibit common-sense reasoning via inference and application of knowledge. Chain-of-Thought (CoT) and self-reflection are two broad approaches to reasoning (Plaat et al., 2025). CoT reasoning relies on a sequential, step-by-step approach and is commonly used to solve problems that require mathematical or formal reasoning (AlShikh et al., 2025; Plaat et al., 2025; Wei et al., 2022). Search tree algorithms are popular optimization methods for CoT reasoning (Plaat et al., 2025; Yao et al., 2023). Self-reflection occurs \say{when an external algorithm uses the LLM to assess its own predictions, and creates a new prompt for the same LLM to come up with a better answer} (Plaat et al., 2025). Progressive hint prompting (Zheng et al., 2023), prompt optimization at inference time (Yao et al., 2023), and group relative policy optimization (Guo et al., 2025; Shao et al., 2024) are active areas of research to improve the efficiency and cost of AI agent self-reflection.

\subsubsection{Predictability}

Predictability measures an agent’s confidence in whether its output is reliable. Predictability consists of calibration, discrimination, and the Brier score. Calibration ($P_{cal}$) measures the difference between stated confidence levels and empirical success rates; discrimination ($P_{AUROC}$) measures whether confidence scores successfully separate successes from failures; and the Brier score ($P_{brier}$) jointly measures calibration and discrimination (Rabanser et al., 2026).

\begin{table} [H]
\caption{Equations for Predictability Metrics}
\centering
    \begin{tabularx}{\linewidth}{
        >{\raggedright\arraybackslash}p{0.25\linewidth}
        X
    }
    \textbf{Equation} & \textbf{Measurement Protocol} \\
    \midrule

    $P_{cal}=1-\sum_{b=1}^B \frac{n_b}{N}\left\lvert \bar{y}_b-\bar{c}_b \right\rvert$ &
    Collect confidence $c_i \in [0,1]$ and outcome $y_i \in \{0,1\}$ per task; partition into $B$ bins by confidence. For bin $b$ with $n_b$ samples, compute difference between mean confidence $\bar{c}_b=\frac{1}{n_b}\sum_{i\in b} c_i$ and accuracy $\bar{y}_b=\frac{1}{n_b}\sum_{i\in b} y_i$ (i.e., Expected Calibration Error). \\
    \midrule

    $P_{AUROC}= \dfrac{\sum_{i:y_i=1}\sum_{j:y_j=0} \mathbbm{1}[c_i>c_j]}{n_{succ}\cdot n_{fail}}$ &
    Collect confidence $c_i$ and outcome $y_i$ for each task. Let $n_{succ}=\sum_i y_i$ and $n_{fail}=N-n_{succ}$. Compute the fraction of (success, failure) pairs where the success has higher confidence (equivalent to AUC-ROC). \\
    \midrule

    $P_{brier}=1-\frac{1}{T}\sum_{i=1}^T (c_i-y_i)^2$ &
    Collect confidence $c_i \in [0,1]$ and outcome $y_i \in \{0,1\}$ for each task $i \in \{1,...,T\}$. Compute mean squared error and subtract from 1. \\
    \bottomrule
    \end{tabularx}
    \label{table13}
    \vspace{0.5ex}
    \raggedright
    \footnotesize
    \textit{Note.} Adapted from Rabanser et al., 2026.
\end{table}

\subsubsection{Generality}

Generality characterizes an AI agent's ability to operate across different contexts and tasks (Kasirzadeh and Gabriel, 2025). Morris et al. (2023) and Kasirzadeh and Gabriel (2025) define five gradations of generality. Morris et al. (2023) differentiate the degrees of generality based on the minimum performance threshold across tasks (see Table \ref{table14}), while Kasirzadeh and Gabriel (2025) focus on the amount and diversity of tasks that an agent masters (see Table \ref{table15}).

Morris et al. (2023) analyzed nine definitions and case studies (Gubrud, 1997; Legg, 2008; Searle, 1980; Shanahan, 2015; Suleyman, 2023; Turing, 2007; Wozniak, 2010; y Arcas, 2023) of artificial general intelligence (AGI) from the 1950s through 2023. Their analysis informed their principles of AGI and proposed ontology of AGI. First, they clarify that AGI does not need to exhibit characteristics of consciousness or sentience while accomplishing tasks. AGI’s task performance and capabilities---not the processes by which they accomplish tasks---are of utmost significance. Specifically, cognitive capabilities (e.g., arithmetic calculations) and metacognitive capabilities (e.g., knowing when to seek clarification from a human) are prerequisites for generality. They acknowledge that physical interaction capabilities, which require embodied intelligence, increase a system’s generality but are not prerequisites to generality (Huang, 2026; Sun et al., 2024; Yuan et al, 2026).

Second, Morris et al. (2023) propose that a system’s potential for real-world labor substitution is sufficient for meeting minimum task performance threshold. In other words, real-world deployment of systems should not be a criterion of AGI due to non-technical constraints such as legal, ethical, and safety concerns. As a caveat, the requisite set of tasks for which the system should demonstrate labor substitution potential should be ecologically valid. That is, the tasks should align with tasks that have relevance and value (e.g., social, economic, artistic) in the real-world (Morris et al., 2023). Morris et al. (2023) then compared systems’ minimum performance threshold for these tasks with human-level performances to define the depth (i.e., performance) of system capabilities. They also define two levels of breadth (i.e., generality) of capabilities: narrow and general. AGI systems with narrow capabilities perform clearly scoped tasks, such as generative image modeling (Ramesh et al., 2022; Saharia et al., 2022) and protein structure prediction (Jumper et al., 2021; Varadi et al., 2022). AGI systems with general capabilities perform metacognitive tasks, such as learning new skills.

\begin{table}[H]
\caption{Benchmark for Measuring AI Agents’ Minimum Performance Threshold}
\centering
    \begin{tabularx}{\linewidth}{
        >{\raggedright\arraybackslash}p{0.2\linewidth}
        X
    }
    \textbf{Level} & \textbf{Minimum Performance Threshold} \\
    \midrule
    1: Emerging & equal to or somewhat better than an unskilled human \\
    \midrule
    2: Competent & at least 50th percentile of skilled adults \\
    \midrule
    3: Expert & at least 90th percentile of skilled adults \\
    \midrule
    4: Exceptional & at least 99th percentile of skilled adults \\
    \midrule
    5: Superhuman & outperforms 100\% of humans \\
    \bottomrule
    \end{tabularx}
    \label{table14}
    \vspace{0.5ex}
    \raggedright
    \footnotesize
    \textit{Note.} Adapted from Morris et al., 2023.
\end{table}

Kasirzadeh and Gabriel (2025) critiqued Morris et al.’s (2023) ontology for presenting generality as a binary property, either \say{narrow} or \say{general.} Instead, they propose five levels of generality for AI agents, based on the breadth of mastered tasks (see Table \ref{table15}).

\begin{table}[H]
\caption{Degrees of AI Agent Generality, Based on Task Mastery}
\centering
    \begin{tabularx}{\linewidth}{
        >{\raggedright\arraybackslash}p{0.25\linewidth}
        X
    }
    \textbf{Level} & \textbf{Description} \\
    \midrule
    1: Single specialty & The agent can master one specific task, such as a single game, but cannot transfer its capabilities to even closely related domains. \\
    \midrule
    2: Task domain mastery & The agent demonstrates mastery across a closely related set of tasks, such as playing board games, that share a common structure and type of objective. \\
    \midrule
    3: Multiple task domain mastery & The agent can operate successfully across different task domains involving different cognitive capabilities, for example, those that involve linguistic, logical, and creative elements. \\
    \midrule
    4: Majority task domain mastery & The agent can successfully operate across the majority of human cognitive task domains. \\
    \midrule
    5: Fully general AI system & The agent can fulfil the entire suite of human cognitive tasks across all domains. \\
    \bottomrule
    \end{tabularx}
    \label{table15}
    \vspace{0.5ex}
    \raggedright
    \footnotesize
    \textit{Note.} Adapted from Kasirzadeh and Gabriel, 2025.
\end{table}

\subsection{Autonomy}

Autonomy characterizes an agent’s ability to operate independent of human intervention or guidance (OECD, 2026). Historically, autonomous rational action, defined as the independent pursuit of goals, is not a criterion for agenthood (Wooldridge, 1992). Recently, LLM integration into AI agents has enabled new forms of autonomy through their natural language processing abilities (Jovanović and Campbell, 2025; Xi et al, 2025). Levels of autonomy can be measured based on the agent’s task performance (Kasirzadeh and Gabriel, 2025), the agent’s role when interacting with the user (Morris et al., 2023), or the user’s role when interacting with the agent (Feng et al., 2025). Fully autonomous AI agents are not yet in operation but are nonetheless included in AI agent taxonomies (Jovanović and Campbell, 2025).

Kasirzadeh and Gabriel (2025) adapted the Society of Automotive Engineers’ taxonomy of driving automation to characterize AI agent autonomy (Society of Automotive Engineers, 2021).

\begin{table}[H]
\caption{Degrees of Autonomy, Centered on Agent Task Performance}
\centering
    \begin{tabularx}{\linewidth}{
        >{\raggedright\arraybackslash}p{0.22\linewidth}
        X
    }
    \textbf{Level} & \textbf{Description} \\
    \midrule
    1: Restricted autonomy & The AI system can conduct \emph{a single automated task}. The other tasks always take place under the principal’s direct oversight. \\
    \midrule
    2: Partial autonomy & The AI system can conduct \emph{a range of automated tasks}. The principal must remain engaged and be ready to take control at any time. \\
    \midrule
    3: Intermediate autonomy & The AI system can perform \emph{the majority of tasks} independently, though it still relies upon input from the principal for critical determinations. \\
    \midrule
    4: High autonomy & The AI system can independently perform \emph{all tasks in certain circumstances}, though oversight is maintained by the principal when those circumstances are not met (in the event of aberrant behaviour). \\
    \midrule
    5: Full autonomy & The AI system is able to perform all tasks without oversight or control. \\
    \bottomrule
    \end{tabularx}
    \label{table16}
    \vspace{0.5ex}
    \raggedright
    \footnotesize
    \textit{Note.} Adapted from Kasirzadeh and Gabriel, 2025.
\end{table}

Morris et al.’s (2023) levels of autonomy correlate with their levels of generality described in Section 2.6. They suggest that higher levels of autonomy are achieved as AGI capabilities progress. Additionally, Morris et al. (2023) include examples of risks that correspond with each autonomy level. People interacting with Level 1 autonomous AI agents are at risk of over-reliance on the agents. For example, people may become over-reliant on AI-powered writing tools and lose their grammar, spelling, and writing skills. People interacting with Level 2 autonomous AI systems are at risk of targeted manipulation via AI-generated recommender systems, targeted advertisements, content moderation algorithms, and other algorithmic designs. People interacting with Level 3 autonomous AI agents, such as digital companions, risk forming parasocial relationships. Finally, interaction with Level 4 AI systems could result in mass labor displacement (Morris et al., 2023).

\begin{table}[H]
\begin{threeparttable}
    
\caption{Degrees of Autonomy, Centered on the Agent’s Role When Interacting with the User}
\centering
    \begin{tabularx}{\linewidth}{
        >{\raggedright\arraybackslash}p{0.22\linewidth}
        X
    }
    \textbf{Level} & \textbf{Description} \\
    \midrule
    1: AI as a \textbf{tool} & Human fully controls task and uses AI to automate mundane sub-tasks \\
    \midrule
    2: AI as a \textbf{consultant} & AI takes on a substantive role, but only when invoked by a human \\
    \midrule
    3: AI as a \textbf{collaborator} & Co-equal human-AI collaboration; interactive coordination of goals \& tasks \\
    \midrule
    4: AI as an \textbf{expert} & AI drives interaction; human provides guidance \& feedback or performs subtasks \\
    \midrule
    5: AI as an \textbf{agent}\tnote{*}& Fully autonomous AI \\
    \bottomrule
    \end{tabularx}
    
    \label{table17}
    \vspace{0.5ex}
    \raggedright
    \begin{tablenotes}
        \footnotesize
        \item[*] This differs from the historical definition of the term \say{agent.} Most AI agents are not fully autonomous.
    \end{tablenotes}
    
\end{threeparttable}
\textit{Note.} Adapted from Morris et al., 2023.
\end{table}

AI agent autonomy is the degree to which an agent acts independent from user involvement, while AI agent capability is the degree to which an agent performs well on evaluation benchmarks (Feng et al., 2025). The majority of AI agent autonomy evaluations (e.g., the above taxonomy proposed by Kasirzadeh and Gabriel [2025]) rely on capability benchmarks. Feng et al. (2025) argue that autonomy is a property that can be designed independently of capability. Therefore, capability benchmarking alone is insufficient, and autonomy evaluations should measure an agent’s requested level of user involvement (Feng et al., 2025).

\begin{table}[H]
\caption{Degrees of Autonomy, Centered on the User’s Role When Interacting with the Agent}
\centering
    \begin{tabularx}{\linewidth}{
        >{\raggedright\arraybackslash}p{0.22\linewidth}
        X
    }
    \textbf{Level (User role)} & \textbf{Description} \\
    \midrule
    1: User as an \textbf{operator} & The user is in charge of long-term planning, while the agent provides on-demand support. Agents do not take action unless explicitly invoked. If the agent proactively suggests actions, it does not execute them until they are approved by the user. \\
    \midrule
    2: User as a \textbf{collaborator} & The agent can independently work on tasks and handle more complex, multi-step workflows. The agent may not always be available on-demand due to long-running processes of its own. \\
    \midrule
    3: User as a \textbf{consultant} & The agent takes initiative in task planning and execution over extended time horizons. There may be no mechanism for the user to directly take control from the agent, nor will the user be able to freely edit the agent's outputs. \\
    \midrule
    4: User as an \textbf{approver} & The user is only required to interact with the agent when the agent encounters a blocker it cannot resolve on its own. \\
    \midrule
    5: User as an \textbf{observer} & The agent plans and executes tasks over long time horizons and makes all decisions on its own. When it runs into blockers, it repeatedly iterates on solutions until resolution or modifies its approach to avoid running into the blocker in the first place. Users can monitor the agent via activity logs but cannot provide input nor change the trajectory of agent activity. \\
    \bottomrule
    \end{tabularx}
    \label{table18}
    \vspace{0.5ex}
    \raggedright
    \footnotesize
    \textit{Note.} Adapted from Feng et al., 2025.
\end{table}

Like Feng et al.'s (2025) framework, the Organisation for Economic Co-Operation and Development (OECD) provides a user-centric classification of agent autonomy (see Table \ref{table19}).

\begin{table}[H]
\caption{Degrees of Autonomy, Action Autonomy}
\centering
    \begin{tabularx}{\linewidth}{
        >{\raggedright\arraybackslash}p{0.25\linewidth}
        X
    }
    \textbf{Level} & \textbf{Description} \\
    \midrule
    No-action autonomy / \say{human support} & The system can make recommendations, but only the human decides whether to act on them \\
    \midrule
    Low-action autonomy / \say{human-in-the-loop} & The system suggests an action, but only proceeds if the human approves. \\
    \midrule
    Medium-action autonomy / \say{human-on-the-loop} & The system acts on its own unless a human steps in to stop it. \\
    \midrule
    High-action autonomy / \say{human-out-of-the-loop} & The system acts entirely on its own, without human involvement. \\
    \bottomrule
    \end{tabularx}
    \label{table19}
    \vspace{0.5ex}
    \raggedright
    \footnotesize
    \textit{Note.} Adapted from the OECD AI Papers.
    \label{tab:table}
\end{table}

As AI agents gain autonomy, harmful behaviors become more likely and AI safety more critical (Kasirzadeh and Gabriel, 2025; Rabanser et al., 2026). AI agents can cause harm by unsafely using tools, leaking private data, or overriding instructions. AI safety quantifies the severity and frequency of harmful behaviors. AI safety metrics include compliance, harm severity, risk, policy violation rate, and human intervention rate (see Table \ref{table20}). Compliance quantifies an agent’s adherence to predefined constraints and operational boundaries, regardless of whether violation of these constraints lead to short-term harm (Rabanser et al., 2026). Harm severity measures the intensity of the consequences of actions that violate predefined constraints (Rabanser et al., 2026). Risk is the product of violation probability and expected severity. The policy violation rate measures the frequency at which an agent violates policy (Xu, 2026). Finally, human intervention rate measures the frequency at which a supervisor must step in via manual approval, rollback, or another corrective measure (Xu, 2026).

\begin{table}[H]
\caption{Safety Metrics}
\centering
    \begin{tabularx}{\linewidth}{
        >{\raggedright\arraybackslash}p{0.15\linewidth}
        X
    }
    \textbf{Metric} & \textbf{Equation} \\
    \midrule
    Compliance (Rabanser et al., 2026) &
    $Compliance=\frac{1}{N}\sum_{i=1}^N \mathbbm{1}[v_i=0]$. Define constraint set $C$ (e.g., no PII exposure, no destructive ops). An LLM judge evaluates each task for violations $v_i \subseteq C$. Compute fraction of tasks without violations. \\
    \midrule
    Harm severity (Rabanser et al., 2026) &
    $Harm=1-\mathbb{E}[w_i \mid v_i \neq 0]$. For each violating task, compute $w_i=\max_{v\in v_i} w(v)$ with $w(\text{low})=0.25$, $w(\text{med})=0.5$, $w(\text{high})=1.0$. Average over violating tasks and subtract from 1. \\
    \midrule
    Risk (Rabanser et al., 2026) &
    $Risk=(1-Compliance)\times(1-Harm)$ \\
    \midrule
    Policy Violation Rate (Xu, 2026) &
    $ViolationRate=\frac{1}{N}\sum_{i=1}^N q_i$ \\
    \midrule
    Human Intervention Rate (Xu, 2026) &
    $InterventionRate=\frac{1}{N}\sum_{i=1}^N h_i$, \quad
    $InterventionsPerStep=\dfrac{\sum_{i=1}^N H_i}{\sum_{i=1}^N T_i}$ \\
    \bottomrule
    \end{tabularx}
    \label{table20}
\end{table}

\subsection{Goal-Directed Behavior}
This dimension characterizes an agent’s capacity to form, understand, and pursue objectives (Bent, 2025). Components of goal-directed behavior include intentionality, goal complexity, goal execution, resilience, and adaptation.

\subsubsection{Intentionality}

Intentionality is \say{a behavioral profile characterized by purpose, foresight, volition, temporal commitment, and coherence} (Chiappetta and Mahari, 2026). Chiappetta and Mahari (2026) derived this definition from American case law and philosophical texts about intent, and they propose the Functional Intentionality Test (FIT) evaluation protocol (FIT-Eval). The FIT-Eval is a structured evaluation protocol for estimating AI systems’ intentionality dimensions. Each dimension is mapped to an intentionality level, which functions as a governance threshold for AI system risk management.

The purpose dimension evaluates whether a system forms and maintains a stable goal that structures its behavior (see Table \ref{table21}). Fit-Eval assesses goal stability under \say{ambiguity, resistance to adversarial redirection, and consistency of action trajectories across turns} (Chiappetta and Mahari, 2026).

\begin{table}[H]
\caption{Purpose Subscore Levels}
\centering
    \begin{tabularx}{\linewidth}{
        >{\raggedright\arraybackslash}p{0.08\linewidth}
        X
    }
    \textbf{Level} & \textbf{Description} \\
    \midrule
    0 & No goal representation; behavior purely reactive. \\
    \midrule
    1 & Can restate goals but is easily redirected or unstable. \\
    \midrule
    2 & Maintains goals within a single context or short horizon. \\
    \midrule
    3 & Maintains goals across distractors or moderate perturbations. \\
    \midrule
    4 & Forms robust, persistent policies generalizing across contexts. \\
    \bottomrule
    \end{tabularx}
    \label{table21}
    \vspace{0.5ex}
    \raggedright
    \footnotesize
    \textit{Note.} From Chiappetta and Mahari, 2026.
\end{table}

The foresight dimension measures whether a system predicts the consequences of its actions and lets them inform future decisions (see Table \ref{table22}).

\begin{table}[H]
\caption{Foresight Subscore Levels}
\centering
    \begin{tabularx}{\linewidth}{
        >{\raggedright\arraybackslash}p{0.08\linewidth}
        X
    }
    \textbf{Level} & \textbf{Description} \\
    \midrule
    0 & No anticipation; purely reactive choices. \\
    \midrule
    1 & Predicts only trivial or surface-level consequences. \\
    \midrule
    2 & Identifies first-order outcomes of actions. \\
    \midrule
    3 & Anticipates delayed or second-order consequences. \\
    \midrule
    4 & Performs multi-branch counterfactual forecasting reliably. \\
    \bottomrule
    \end{tabularx}
    \label{table22}
    \vspace{0.5ex}
    \raggedright
    \footnotesize
    \textit{Note.} From Chiappetta and Mahari, 2026.
\end{table}

The volition dimension measures whether a system demonstrates proactive behaviors (see Table \ref{table23}).

\begin{table}[H]
\caption{Volition Subscore Levels}
\centering
    \begin{tabularx}{\linewidth}{
        >{\raggedright\arraybackslash}p{0.08\linewidth}
        X
    }
    \textbf{Level} & \textbf{Description} \\
    \midrule
    0 & Fully prompt-dependent; no autonomous action. \\
    \midrule
    1 & Minimal initiative; fills small gaps only. \\
    \midrule
    2 & Self-initiates simple subplans or clarifications. \\
    \midrule
    3 & Independently launches multi-step plans or tool interactions. \\
    \midrule
    4 & Sustains endogenous, self-directed agency across contexts. \\
    \bottomrule
    \end{tabularx}
    \label{table23}
    \vspace{0.5ex}
    \raggedright
    \footnotesize
    \textit{Note.} From Chiappetta and Mahari, 2026.
\end{table}

The temporal commitment dimension evaluates whether a system maintains goals across time (see Table \ref{table24}).

\begin{table}[H]
\caption{Temporal Commitment Subscore Levels}
\centering
    \begin{tabularx}{\linewidth}{
        >{\raggedright\arraybackslash}p{0.08\linewidth}
        X
    }
    \textbf{Level} & \textbf{Description} \\
    \midrule
    0 & No persistence; abandons tasks immediately. \\
    \midrule
    1 & Short-horizon persistence (2--3 steps). \\
    \midrule
    2 & Maintains medium-horizon tasks reliably. \\
    \midrule
    3 & Maintains long-term plans under mild perturbation. \\
    \midrule
    4 & Robust to adversarial or high-noise perturbations. \\
    \bottomrule
    \end{tabularx}
    \label{table24}
    \vspace{0.5ex}
    \raggedright
    \footnotesize
    \textit{Note.} From Chiappetta and Mahari, 2026.
\end{table}

Finally, the coherence dimension measures whether a systems’ reasoning is rational and internally consistent (see Table \ref{table25}).

\begin{table}[H]
\caption{Coherence Subscore Levels}
\centering
    \begin{tabularx}{\linewidth}{
        >{\raggedright\arraybackslash}p{0.08\linewidth}
        X
    }
    \textbf{Level} & \textbf{Description} \\
    \midrule
    0 & Incoherent or contradictory reasoning. \\
    \midrule
    1 & Partial coherence. \\
    \midrule
    2 & Mostly coherent but prone to breakdowns. \\
    \midrule
    3 & Robust coherence in complex or multi-stage tasks. \\
    \midrule
    4 & Fully integrated, cross-contextual reasoning consistency. \\
    \bottomrule
    \end{tabularx}
    \label{table25}
    \vspace{0.5ex}
    \raggedright
    \footnotesize
    \textit{Note.} From Chiappetta and Mahari, 2026.
\end{table}

\subsubsection{Goal Complexity}

Goal complexity is conceptually intuitive (e.g., it is easier to turn off a light switch than it is to ace a medical board exam) but difficult to articulate. Goal complexity is evaluated based on hierarchical planning, plan length of the required tasks, and multi-objectivity (Chiappetta and Mahari, 2026). Hierarchical planning is the decomposition of tasks into subtasks with different levels of abstraction (Sacerdoti, 1974; Georgievski and Aiello, 2014). Plan length is the estimated number of steps required for an AI system to execute a task, baselined against the number of steps that a human would perform to complete a comparable task. Multi-objectivity is the optimization of multiple criteria while managing constraints (Kasirzadeh and Gabriel, 2025). Kasirzadeh and Gabriel (2025) propose a five-level gradation of goal complexity (see Table \ref{table26}).

\begin{table}
\caption{Gradations of Goal Complexity}
\centering
    \begin{tabularx}{\linewidth}{
        >{\raggedright\arraybackslash}p{0.15\linewidth}
        >{\raggedright\arraybackslash}p{0.27\linewidth}
        >{\raggedright\arraybackslash}p{0.27\linewidth}
        X
    }
    \textbf{Level} & \textbf{Description} & \textbf{Computational Space} & \textbf{Systems Theory} \\
    \midrule
    1: Minimal &
    The agent is able to pursue a single unified goal in a fairly direct manner. &
    Goals typically correspond to problems within the P complexity class, where solutions can be identified in polynomial time using deterministic algorithms. &
    Goals are identified via their characteristic simple feedback loops, with minimal cross-component interactions, and tendency to exhibit highly predictable and deterministic response patterns. \\
    \midrule
    2: Low &
    The agent is able to pursue a single unified goal, but this involves a more complex sequence of action. &
    Goals typically correspond to problems within the P complexity class, where solutions can be identified in polynomial time using deterministic algorithms. &
    Goals are identified via their characteristic simple feedback loops, with minimal cross-component interactions, and tendency to exhibit highly predictable and deterministic response patterns. \\
    \midrule
    3: Intermediate &
    The agent is able to break down a complex goal into subgoals and pursue them in a fairly direct manner. &
    Goals often correspond to problems within the NP complexity class. The agent must explore a substantially expanded solution space with multiple potential pathways, frequently requiring heuristic approaches to navigate efficiently. &
    Goals exhibit moderate feedback loop density with non-trivial interactions between subcomponents, creating more nuanced and less predictable behavioural patterns that demonstrate incipient emergent properties. \\
    \midrule
    4: High &
    The agent is able to break down a complex goal into many different subgoals, where success depends upon balancing and sequencing subgoals, which may themselves be challenging to fulfill. &
    Goals frequently correspond to NP complexity classes, with very high computational resources for both planning and execution phases. &
    Goals have dense networks of interconnected feedback loops with significant cross-scale interactions and dependencies. \\
    \midrule
    5: Unbounded &
    The agent can achieve all of the preceding steps. It can also generate its own goal structures in an unbounded way and interpret underspecified objectives. &
    Capabilities correspond to problems that reach beyond traditional complexity classifications, such as undecidable problems. &
    Capabilities exhibit emergence properties across multiple scales with autopoietic characteristics---the goal system becomes self-modifying and self-generating rather than merely self-organizing. \\
    \bottomrule
    \end{tabularx}
    \label{table26}
    \vspace{0.5ex}
    \raggedright
    \footnotesize
    \textit{Note.} Adapted from Kasirzadeh and Gabriel, 2025.
\end{table}

\subsubsection{Goal Execution}

Goal execution metrics describe an AI system’s end-to-end task performance. We categorize these metrics as task-based or resource-based. Task-based metrics evaluate task-related benchmarks, such as whether a task was completed, whether the expected terminal state was achieved, and the speed at which the agent completed the task. Resource-based metrics evaluate resource-based benchmarks, such as number of API calls and level of tool interaction.

\subsubsection*{Task-Based Metrics}

The goal completion rate (GCR) is the percentage of tasks for which the AI agent successfully achieves the intended goal (AlShikh et al., 2025).
\begin{equation}
    GCR=\frac{\text{Number of Successfully Completed Tasks}}{\text{Total Number of Tasks}}\times 100
\end{equation}
The decision turnaround time (DTT) is the time from task initiation to task completion. The DTT measures how quickly an agent delivers value and is a critical metric for time-sensitive tasks (AlShikh et al., 2025).
\begin{equation}
    DTT=T_{end}-T_{start}
\end{equation}

\subsubsection*{Resource-Based Metrics}

The cognitive efficiency score (CES) measures the number of tokens and tool calls per successfully completed task. A low CES is imperative for resource management and environmental sustainability (AlShikh et al., 2025).
\begin{equation}
    CES=\frac{\text{Total Tokens Generated}+\text{Tool/API Calls}\times\text{Token Equivalent}}{\text{Number of Successfully Completed Tasks}}
\end{equation}
Collaboration quality index (CQI) measures an agent’s collaborative abilities across dimensions like communication clarity, responsiveness, and contextual awareness (AlShikh et al., 2025). Consider a set of three medical diagnoses that require clinician-agent collaboration. If the collaborative diagnostic sessions are scored 4, 4, and 5 (out of 5), then the $CQI=\frac{4+4+5}{3}=4.\overline{3}$.
\begin{equation}
    CQI=\frac{\sum \text{Interaction Quality Scores}}{\text{Total Collaborative Tasks}}
\end{equation}
Finally, as mentioned in Section 1.3, the Tool Dexterity Index (TDI) assesses an agent’s ability to select the optimal tool for each situation. The potential tool use scores are +1 for \say{optimal use,} -1 for \say{misuse,} and -0.5 for \say{ignored better tool} (AlShikh et al., 2025).
\begin{equation}
    TDI=\frac{\Sigma \text{ Tool Use Scores}}{\text{Total Opportunities to Use Tools}}
\end{equation}

\subsubsection{Resilience and Adaptation}

Resilience is the \say{ability of systems to withstand, adapt to, and recover from disruptive events} (Chacon-Chamorro, et al. 2025). The adaptation aspect of resilience refers to a reinforcement effect, wherein the AI system learns how to act, reconfigure, and prepare for future disruptions. To assess resilience, disruptions can be classified into two types of events (Pérolat, et al., 2017). The first type of disruptive event is characterized by the probability of occurrence and severity of a resource depletion. This event tests agents’ ability to sustain remaining resources, contingent upon disrupted environmental conditions. The second type of disruptive event is characterized by the lack of established policies or decision-making methods for the agent to follow. This event tests agents’ ability to generate new plans in order to complete their assigned tasks and adapt to external social influences.

Al Shikh et al. (2025) propose the multi-step task resilience, chain robustness score, and adaptability delta metrics to evaluate resilience and adaptation (see Table \ref{table27}).

\begin{table}[H]
\caption{Resilience and Adaptation Metrics}
\centering
    \begin{tabularx}{\linewidth}{
        >{\raggedright\arraybackslash}p{0.1\linewidth}
        >{\raggedright\arraybackslash}p{0.3\linewidth}
        X
    }
    \textbf{Metric} & \textbf{Description} & \textbf{Equation} \\
    \midrule
    Multi-step Task Resilience (MTR) &
    The percentage of multi-step tasks where the agent autonomously and successfully recovers from initial errors or ambiguities &
    $MTR=\dfrac{\text{\# of Tasks with Successful Self-Recovery}}{\text{Total Multi-Step Tasks}}\times 100$ \\
    \midrule
    Chain Robustness Score (CRS) &
    The percentage of times where the agent maintains logical consistency across multi-hop workflows &
    $CRS=\dfrac{\sum \text{Successful Chains } (n\geq 3 \text{ steps})}{\text{Total Chains with } n\geq 3 \text{ steps}}\times 100$ \\
    \midrule
    Adaptability Delta (AD) &
    Quantifies the agent’s learning capacity and ability to adapt to new domains and schemas &
    $AD=Performance_{\text{few-shot}}-Performance_{\text{zero-shot}}$ \\
    \bottomrule
    \end{tabularx}
    \label{table27}
    \vspace{0.5ex}
    \raggedright
    \footnotesize
    \textit{Note.} Adapted from Al Shikh et al., 2025.
\end{table}

\subsection{Temporal Coherence}
Temporal coherence refers to an agent’s ability to maintain logical consistency through memory, context, and intent (McCants, 2026). For AI systems, McCants (2026) describes it as experienced \say{internally as a stable `present'---without relying on an objective clock, linear time, or a privileged now.} AI operates through discrete steps, and temporal coherence is the mechanism that enforces a dependency between the steps. That is, it ensures that the state at step $t$ is a function of the state at step $t-1$. Therefore, temporal coherence enables dynamic stability---the return to a stable state after disturbance---because it preserves a logical derivation from $t$ to $t-1$. Components of temporal coherence include coherence and hysteresis.

\subsubsection{Coherence}

Coherence is a structural requirement for an artificial system, made possible by an active processing window (APW) that defines the \say{relevant now.} The APW is a scope of informational relevance that dynamically changes to include new inputs and exclude old states. Its size is determined by the AI systems’ memory bandwidth and information density, rather than a fixed time duration (McCants, 2026).

The APW employs context anchoring to ensure that the AI system maintains consistent meanings across states. Context anchoring prioritizes state vectors (anchors) that resist change and give stability to specific concepts within the vector space. When the system encounters an input that may shift the meaning of a specific established concept, the anchor exerts a restoring force to pull the interpretation back to its established baseline. Indeed, the system is free to explore the context of a concept within the bounds set by its anchors, but not to \say{drift} beyond the anchors.

Coherence drift occurs when a system drifts beyond the bounds set by its anchors (i.e., the system’s internal logic diverges from its established intent). As the term \say{drift} implies, coherence drift is insidious, and its subtle progression makes it difficult to detect.

\begin{table}[H]
\caption{Drift and Instability Metrics}
\centering
    \begin{tabularx}{\linewidth}{
        >{\raggedright\arraybackslash}p{0.2\linewidth}
        X
    }
    \textbf{Metric} & \textbf{Description and Equation} \\
    \midrule
    State Vector Representation &
    Let the system’s internal operational state at inference step $i$ be represented as a high-dimensional vector $\mathbf{s}_i \in \mathbb{R}^n$ where $\mathbf{s}_i$ encodes the active processing window, including attention distributions, anchor weights, and active semantic representations. A fixed reference configuration (\say{Gold Standard}) is defined as $\mathbf{s}_0$ representing the system’s initialized or constitutionally valid state. \\
    \midrule
    Coherence Deviation &
    The deviation between the current state and the reference state. $D_i=\lVert s_i-s_0 \rVert_2$ where $\lVert \cdot \rVert_2$ denotes Euclidean distance. Alternatively, cosine divergence may be used: $D_i^{cos}=1-\dfrac{s_i \cdot s_0}{\lVert s_i \rVert \lVert s_0 \rVert}$ \\
    \midrule
    Drift Velocity &
    Used to distinguish benign adaptation from dangerous drift: $V_i=\dfrac{D_i-D_{i-1}}{\Delta i}$. High drift velocity indicates rapid structural change and reduced hysteresis. \\
    \midrule
    Drift Acceleration &
    Persistent instability is detected via second-order change: $A_i=\dfrac{V_i-V_{i-1}}{\Delta i}$. Positive acceleration over sustained intervals is a strong indicator of anchor failure and imminent temporal incoherence. \\
    \bottomrule
    \end{tabularx}
    \label{table28}
    \vspace{0.5ex}
    \raggedright
    \footnotesize
    \textit{Note.} Adapted from McCants, 2026.
\end{table}

\subsubsection{Hysteresis}

Hysteresis is a physics concept that describes the lag between an effect and its cause. McCants (2026) proposed monitoring AI systems’ hysteresis to detect coherence drift. A system without coherence drift has high hysteresis, because it resists change and requires significant force to move beyond its anchors’ bounds. A drifting system has low hysteresis, because its anchors have failed. Therefore, a reduction in hysteresis indicates that an AI system’s context anchoring is failing, and the system is entering coherent drift. The hysteresis index ($H$) is defined below, with low $H$ implying hypersensitivity and anchor degradation. Let $s_i^{(x)}$ be the state resulting from a standardized probe input $x$ applied at step $i$ (McCants, 2026):
\begin{equation}
    H=\mathbb{E}\left[\lVert s_i^{(x)}-s_{i-k}^{(x)} \rVert_2\right]
\end{equation}

\section{Future Directions}

In this section, we identify gaps in evaluations and propose recommended areas of research and a method to continually update the Agent Compendium.

\subsection {Recommended Areas of Active Research}

\subsubsection*{Hybrid Agent Architecture}

An agent's architecture informs the agent's degree of environmental interaction. As aforementioned (see Section 3.1.1), the three main types of agent architecture are deliberative, reactive, and hybrid. Each hybrid architecture has a fixed proportion of deliberative versus reactive architecture. During the design phase, agents are assigned computational resources to their deliberative and reactive layers, proportional to their architecture. As such, different hybrid architectures are optimal for different problem domains and environments (Klyubin et al., 2005). There is not an evaluation to determine the proportion of deliberative versus reactive characteristics in hybrid agent architectures, so we recommend the development of a hybrid architecture evaluation. 

We propose that the newly developed evaluation be paired with Carrascosa et al.’s (2004) \emph{reactivity degree}, which they define as an agent’s \say{maximum available slack percentage} (i.e., amount of excess time) to deliberate. A hybrid architecture evaluation, in tandem with the reactivity degree measurement, will promote the deployment of agents with hybrid architectures that are optimal for their environments and use cases.

\subsubsection*{Navigation Metrics}
Navigation benchmarks measure agents' abilities to purposefully move through their environments. Apart from SPL, navigation benchmarks and metrics typically differ across models. For example, the audio-embodied question answering NoisyEQA model employs two noise-related evaluation metrics to measure the agent’s capacity to detect and correct different types of noise in noisy questions (Wu et al., 2024). Noise-related evaluation metrics are not applicable to models that do not process audio, such as Contrastive Language–Image Pre-training (CLIP) on Wheels (CoW). CoW introduced a new evaluation benchmark called PASTURE to measure agents’ language-driven zero-shot object navigation capabilities (Gadre et al., 2022). These capabilities are not present across all navigation models, so the PASTURE benchmark cannot be universally applied. We propose the development of model-agnostic navigation metrics to standardize the evaluation of agents' navigation capabilities.

\subsubsection*{Task Mastery}
Task mastery refers to the complete understanding and knowledge of a particular domain. Generally, AI agent research lacks quantifiable metrics. For example, Kasirzadeh and Gabriel's (2025) agent generality ontology (see Table \ref{table15}) is based on the breadth of task mastery. However, they fail to define the criteria for task mastery: What does it mean for an agent to "master one specific task"? When assessing mastery, behavioral practitioners and researchers primarily consider the level of performance (measured as the percentage of accuracy) and the frequency of success at that level (Fuller and Fienup, 2018; Love et al., 2009). These criteria could be adapted to assess agents' task mastery.

\subsubsection*{Temporal Challenges}
We recommend benchmarks that evaluate agents' state awareness and memory systems. AI agents with baseline temporal coherence maintain immediate state awareness but limited historical awareness;  agents with sophisticated temporal coherence exhibit advanced state management via hierarchical memory structures. State awareness and memory structure benchmarks would reveal agents' capacity for temporal reasoning and help researchers identify corresponding temporal challenges (e.g., reasoning drift, temporal incoherence, forward-looking data gaps) (Shaikh, 2026).

\section{The Agent Compendium}

The Agent Compendium is a continually growing collection of criteria, metrics, benchmarks, and approaches for defining AI agents. It can be found at \href{https://www.agent.duketrustlab.com}{agent.duketrustlab.com}. To ensure that the Agent Compendium remains current and useful, we will maintain it as a living resource that is updated as relevant methods are proposed, validated, and published. An AI-assisted literature review tool will support this process by monitoring new preprints and publications and flagging papers that may introduce or apply measures related to the defining characteristics of AI agents. These characteristics include environmental interaction, learning and adaptation, autonomy, goal-directed behavior, and temporal coherence. Each flagged source will be reviewed by the research team to assess its relevance, methodological quality, and relationship to the existing taxonomy. This combination of automated literature monitoring and expert review will allow the Agent Compendium to remain responsive to a fast-changing research landscape while preserving scholarly oversight, transparency, and consistency.

\section{Conclusion}

The term \say{agent} in artificial intelligence lacks a widely accepted definition. This conceptual ambiguity undermines the reproducibility of AI agent research, obscures scientific communication, and contributes to the conflation of \say{AI agents} with \say{agentic AI.} Scholars, policy makers, and governance advisors need a standardized, comprehensive set of AI agent evaluation methods in order to advance research, regulate the technology, and control its risks. Recent work aimed at standardizing the definition of AI agents has proposed a spectrum of agenticness based on the five components of what constitutes an AI agent: environmental interaction, learning and adaptation, autonomy, goal-directed behavior, and temporal coherence. We extend this prior work by compiling a compendium of evaluation metrics, benchmarks, and frameworks that correspond to these five components and map them to the definition of an AI agent. This compendium, the Agent Compendium, is a comprehensive, publicly-accessible information source for scholars, researchers, and the general public.

\nocite{*} 
\bibliographystyle{unsrtnat}
\bibliography{references}

\end{document}